\documentclass[conference]{IEEEtran}
\usepackage{float}

\ifCLASSINFOpdf
  \usepackage[pdftex]{graphicx}
  \DeclareGraphicsExtensions{.pdf,.jpeg,.png}
\else
  \usepackage[dvips]{graphicx}
  \DeclareGraphicsExtensions{.eps}
\fi
\usepackage[cmex10]{amsmath}
\begin{document}
%
\title{On-Device Information Extraction from SMS using Hybrid Hierarchical Classification}

\author{
\IEEEauthorblockN{Shubham Vatsal, Naresh Purre, Sukumar Moharana, Gopi Ramena, Debi Prasanna Mohanty}
\IEEEauthorblockA{Samsung R\&D Institute Bangalore, India\\
\{shubham.v30, naresh.purre, msukumar, gopi.ramena, debi.m\}@samsung.com}
}


%


\maketitle

\begin{abstract}
Cluttering of SMS inbox is one of the serious problems that users today face in the digital world where every online login, transaction, along with promotions generate multiple SMS. This problem not only prevents users from searching and navigating messages efficiently but often results in users missing out the relevant information associated with the corresponding SMS like offer codes, payment reminders etc. In this paper, we propose a unique architecture to organize and extract the appropriate information from SMS and further display it in an intuitive template. In the proposed architecture, we use a Hybrid Hierarchical Long Short Term Memory (LSTM)-Convolutional Neural Network (CNN) to categorize SMS into multiple classes followed by a set of entity parsers used to extract the relevant information from the classified message. The architecture using its preprocessing techniques not only takes into account the enormous variations observed in SMS data but also makes it efficient for its on-device (mobile phone) functionalities in terms of inference timing and size.
\end{abstract}


%
\IEEEpeerreviewmaketitle

\section{Introduction}
Short Message Service (SMS) was the most widely used communication application in the past few years, with an estimated 3.5 billion active users, or about 80\% of all mobile subscribers. Although with the advent of many user interactive messaging applications, some people might feel that it is too late to have any kind of innovative work in an age-old technology like SMS, but statistics tell us a different story. According to SMS marketing trends and statistics of 2018 as reported by BroadNet\footnote{BroadNet. 2018. SMS Marketing Trends and Statistics. (August 2018) \url{https://www.broadnet.me/blog/sms-marketing-trends-and-statistics-dash-2018}}, people read 97\% of SMS and their response rate is around 48\%, which is only 5\% to 6\% in case of email. There are approximately 500 million messages sent in a day that is around 182 billion in a year globally. Extracting entities is a big challenge especially in case of SMS since these SMS do not have a well-defined grammar and most of the information that actually grabs a user’s attention is present in a small phrase rather than the entire SMS text.

Every user's inbox today is flooded with SMS in an unorganized way. This prevents the user from efficiently extracting information from text data present in their SMS inbox. Also, businesses still see SMS as a powerful marketing tool having highest click through rate (CTR) of 90\%, thus guaranteeing that number of marketing SMS are not going to go down anywhere in near future. Manually scrolling through individual SMS to find a month old flight ticket or a coupon which is about to expire is something which really does not belong to this present technologically perceptive world. We introduce a novel pipeline architecture for efficient organization and content presentation of SMS. This entire architecture is designed in a way to make it efficient for its on-device functionalities. The comparison shows that our solution is easily able to outperform the present popular applications when classification of SMS is concerned and along with that, the end content presentation guarantees a major user experience upgrade.

\section{Related Work} \label{sec:related}
There have been many works on information extraction from email, sms and social media text \cite{bontcheva2013twitie,kushmerick2001information}. Most of the previous papers have concentrated only on binary classification of SMS i.e. spam vs ham \cite{almeida2011contributions,cormack2007spam} whereas we take into consideration 18 classes for classification and corresponding information extraction. However, in some papers, they have expanded beyond binary classification of SMS data, but still they have restricted themselves to 4-5 classes and also the size of the dataset that they have considered is less. In \cite{dewi2017multiclass} the authors have expanded binary classification of SMS to multiclass classification by considering four different classes. However, the paper has restricted itself to a small dataset. Small dataset accompanied only with four classes allows shallow learning techniques to give impressive results whereas our corpus size is around 8k consisting of more than 300 different vendors and hence their corresponding templates. In \cite{cooper2005extracting} the author concentrates on information extraction from SMS using pattern-based approach. Pattern based approaches are less efficient for on-device purposes than machine learning approaches.

To the best of our knowledge, there is hardly any work specifically focusing on extracting information from SMS data using text classification techniques. SMS data being one of its kind involves various types of difficulties during its processing. These could be anything ranging from ambiguity associated with the label of a particular SMS to the huge amount of variations involved in extracting the corresponding information from the SMS. In this paper, we present a novel pipeline which efficiently classifies and extracts relevant information from SMS data.

\section{Pipeline Description} \label{sec:pipeline}
This section broadly describes the functionality of individual components and finally talks about the way these components merge into a single pipeline to achieve the task of information extraction from SMS data.

\subsection{Corpus Details} \label{sec:corp}
Firstly, a corpus of around 8k SMS was collected and manually annotated. We built an android application to collect SMS data from around 150 volunteers. We chose volunteers based on the following factors: 1. Demography, 2. Profession, 3. Interests. The above mentioned steps were taken in order to prevent any kind of bias in the data with respect to any of the above mentioned factors. We used a privacy filter to further guarantee the anonymity of the user. This privacy filter performed four defined tasks: 1. Jumble all the numbers, 2. Replace names with random names, 3. Replace URLs with tags, 4. Ignore personal SMS.

\begin{table}[h]
  \caption{Major Class Corpus Statistics}
  \label{tab:corpus_major}
  \centering
  \begin{tabular}{cc}
  \hline
    \bfseries SMS Major Class & \bfseries Number of SMS\\
  \hline
    Info & 1591\\
    Reminder & 2211\\
    Offer & 2801\\
    Transaction & 921\\
    Otp & 854\\
  \hline
  \end{tabular}
\end{table}

We also gave the liberty to volunteers to manually remove any other messages which they were not willing to share. We collected approximately 700-800 SMS from each user. We had a panel of 5 experts for manual annotation of data. We used kappa measure \cite{eugenio2004kappa} to calculate the inter annotator agreement. Our inter annotator agreement was found to be as high as 0.98 as depending on the categories defined by us, the element of ambiguity was inherently less. Tables~\ref{tab:corpus_major} and~\ref{tab:corpus_sub} capture the details of data used for our model at the corresponding two levels of hierarchy. The first level is referred to as “Major Class” level whereas the second level or sub-level is referred as “Sub Class” level.

\begin{table}[h]
  \caption{Sub Class Corpus Statistics}
  \label{tab:corpus_sub}
  \centering
  \begin{tabular}{cc}
  \hline
    \bfseries SMS Sub Class & \bfseries Number of SMS\\
  \hline
    Reminder\_Appointment & 169 \\
    Reminder\_Movie & 101 \\
    Reminder\_Bus & 331 \\
    Reminder\_Train & 106 \\
    Reminder\_Flight & 229 \\
    Reminder\_Bill & 711 \\
    Reminder\_Delivery & 349 \\
    Reminder\_Others & 215 \\
    Offer\_Flight & 225 \\
    Offer\_Shopping & 963 \\
    Offer\_Cab & 215 \\
    Offer\_Food & 393 \\
    Offer\_Hotel & 177 \\
    Offer\_Movie & 184 \\
    Offer\_Others & 644 \\
  \hline
  \end{tabular}
\end{table}

\subsection{Data Preprocessing}
We use a set of techniques to convert a message into a more generalized format, which in return helped us in training a deep learning model on a very small dataset of 8k, and still achieve state of the art results. Smaller dataset also allowed us to have a smaller embedding layer and hence obtain a better performance on-device. We used primarily three types of techniques to achieve the level of abstraction resulting in better results.

\subsubsection{Placeholder Selection}
This technique is used to replace certain type of keywords/patterns in the SMS dataset with placeholders. We replace phone numbers, date and time, currencies, URLs and other numbers in our dataset.

\subsubsection{Removal of Less Frequent Words}
This technique involves removing very less frequent words from the dataset. Low frequency words generally comprise of locality names, person names etc. that does not serve any useful purpose in learning of the model.

\subsubsection{Removal of City Names}
We use a custom dictionary of cities to remove the city names from the dataset as they also have little significance when it comes to learning of the model.

\subsection{Custom Embedding Layer} \label{sec:emb}
There have been many works done in the domain of word embeddings to resolve natural language processing problem statements especially for English language. The reason why we opted for custom word embeddings and not for one of GloVe \cite{pennington2014glove} or other popular embeddings is because SMS data does not adhere to proper grammar which we might expect in case of a news article. In addition, there are many application names, which contribute significantly towards classification of SMS, which cannot be found in other pre-trained word embeddings as they were not trained on SMS dataset.

\begin{figure}
  \centering
  \includegraphics[width=\linewidth]{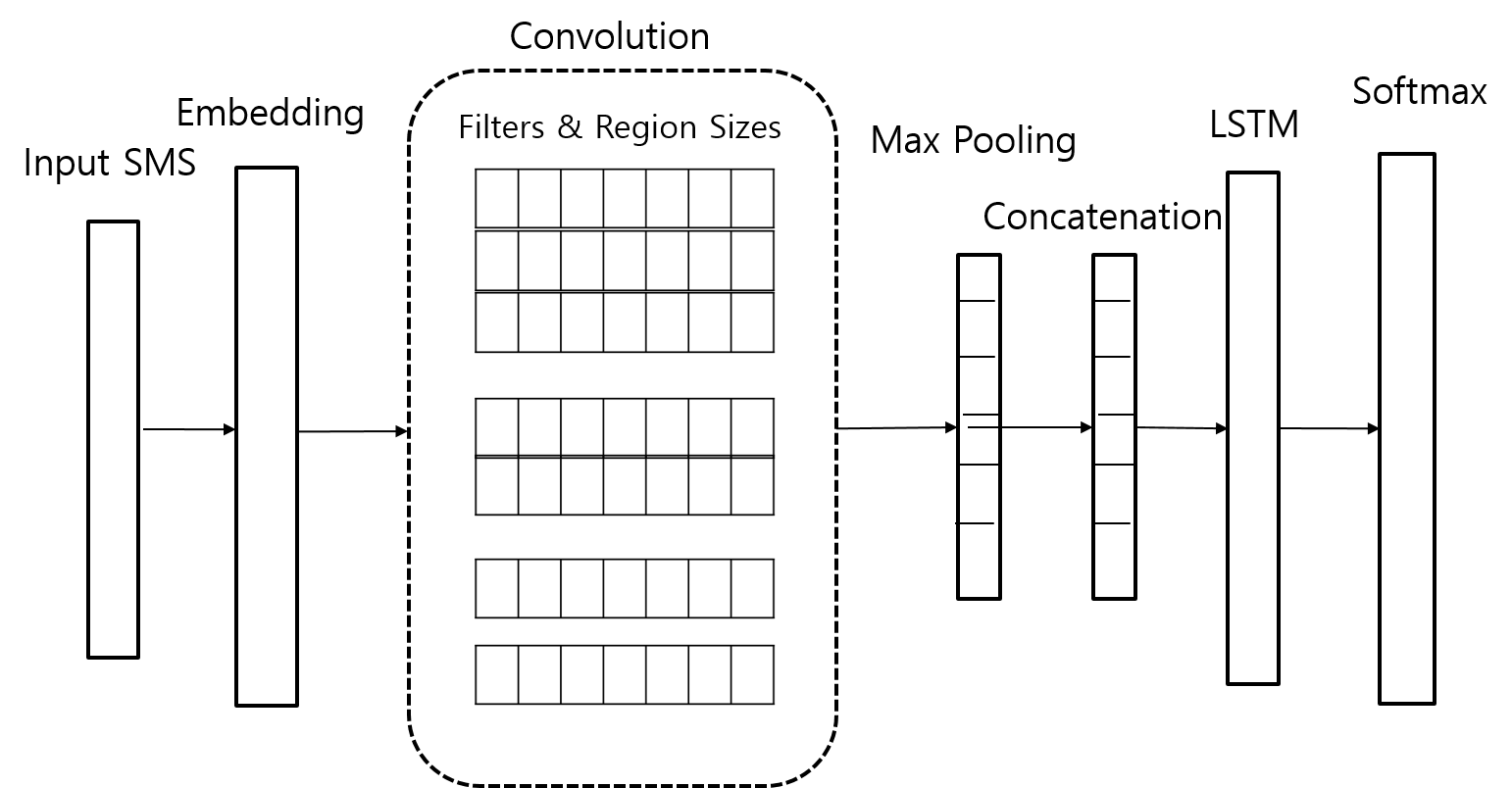}
  \caption{CNN-LSTM Neural Network Architecture}
  \label{fig:model}
\end{figure}

\subsection{Hybrid Hierarchical CNN-LSTM}
In our pipeline, a Hybrid CNN-LSTM model is used to classify the message in any one of the 18 possible categories. It uses a hierarchical approach to achieve the above task. The hierarchical approach uses two levels of classification. First level (Major Class level) classifies the message in one of the five major categories: Info, Offer, Reminder, Transaction or Otp. If the SMS is categorized either as Offer or as Reminder, then we move to second level of classification where the SMS is classified into the corresponding sub classes of Offer and Reminder major category. However, for other major categories, we do not move to second level of classification.

Figure~\ref{fig:model} captures the entire flow of the classification task. The input is first passed to the custom embedding layer, which is created during the backpropagation of learning and has been already discussed in Section~\ref{sec:emb}. Next, the input from custom embedding layer goes to the Convolution layer where the convolution is applied between SMS matrix and Region across all the filters. After the Convolution layer, we have Max Pooling layer which down-samples the input and helps us in getting the most important features of the input. The Concatenation layer joins the received input along a common dimension. For first level classification the output of Concatenation layer goes to an LSTM layer which is further directed to Softmax layer whereas in case sub-level classification the output of Concatenation layer goes directly to Softmax layer. The Softmax layer gives us the probability across all the labels.

\begin{figure}
  \centering
  \includegraphics[width=\linewidth]{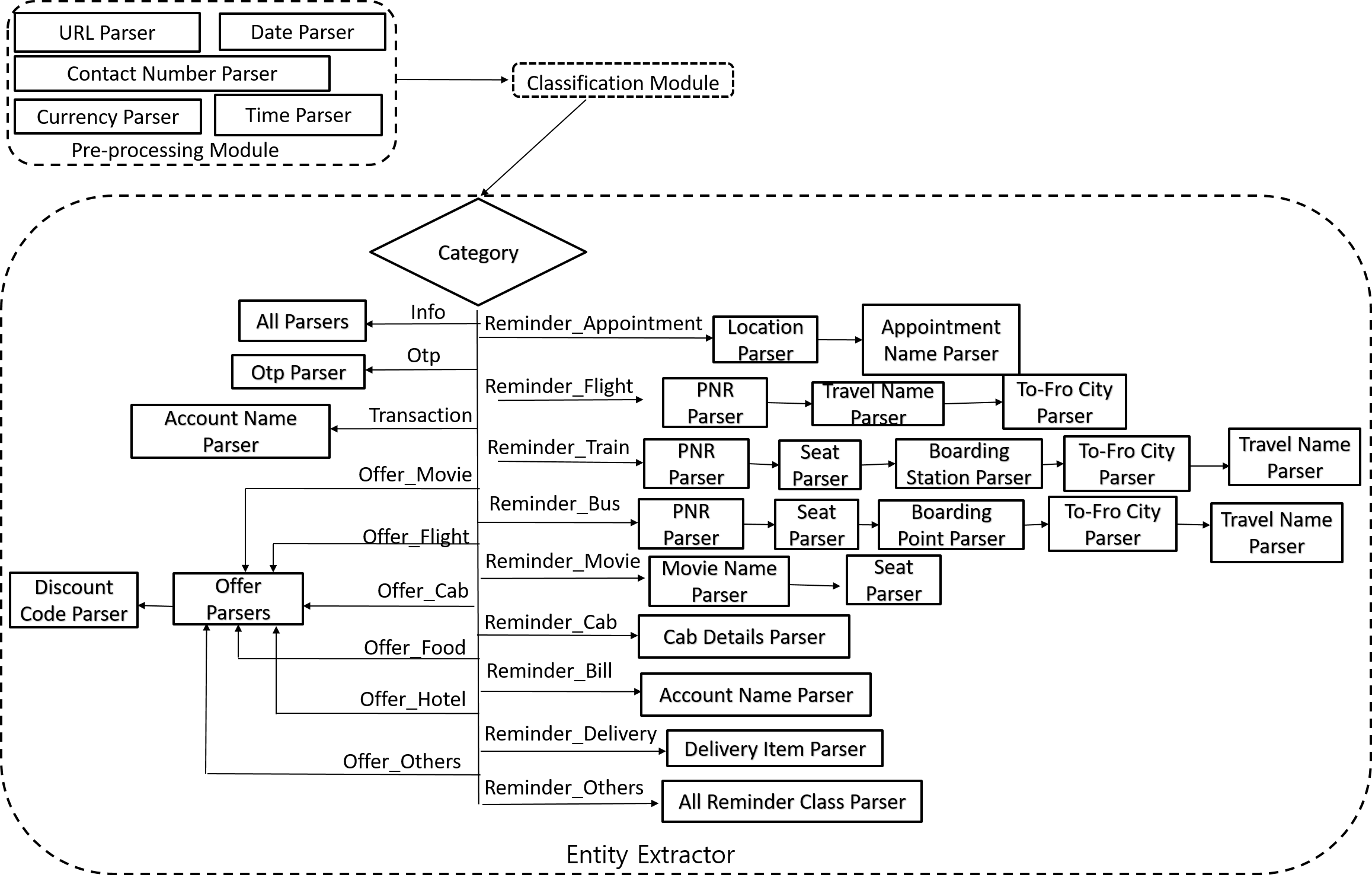}
  \caption{Entity Extractor}
  \label{fig:entext}
\end{figure}

The rationale behind using CNN-LSTM in first level is to handle SMS of the type having two contextual information. These types of SMS have one part indicating some kind of Reminder information whereas the other part of SMS indicates some kind of Offer information. For example, the message "Please pay the due amount of Rs.97 by 3rd May. You can use Paytm to avail a discount of 9\%", the human mind is bound to get confused as the first part of the message indicates a reminder while the second part conveys an offer. Since CNN concentrates on a group of nearby words in its learning phase, it becomes impossible for CNN to correctly identify the main subject of a message having multiple contexts and hence mostly ends up in wrong categorization. Once a Major Class category is determined as Reminder or Offer, CNN can easily identify the corresponding Sub Class categories.The hyperparameter values for the CNN-LSTM model are listed in Table~\ref{tab:hyperparameters}.

\begin{table}[h]
  \caption{Hyperparameters of Hybrid Model}
  \label{tab:hyperparameters}
  \centering
  \begin{tabular}{cc}
  \hline
    \bfseries Hyperparameter & \bfseries Value\\
  \hline
    Embedding Dimension & 128 \\
    Region Sizes & 2,3 \\
    Number of Filters & 128 \\
    Batch Size, Epochs, Dropout & 16, 200, 0.6 \\
    Number of hidden LSTM units & 120 \\
    Development Dataset & 0.1 of Training Set \\
  \hline
  \end{tabular}
\end{table}

\subsection{Entity Extractor}
Entity Extractor is used to extract a set of entities present in a message, which are further used in appropriate rendering of SMS. There are a certain set of parsers as shown in the preprocessing module in Figure~\ref{fig:entext}, which are applied irrespective of the category predicted because these information can exist in any category of SMS. Apart from these parsers, since category Info is the fallback category in Major Level classification, it can contain voluminous amount of variations, thus employing all kinds of parsers for information extraction.

\begin{figure}[h]
  \centering
  \includegraphics[width=0.9\linewidth]{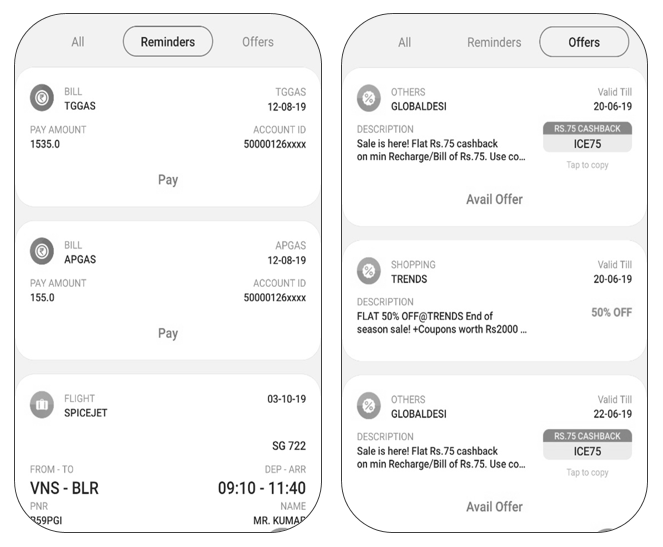}
  \caption{Content Presentation}
  \label{fig:content}
\end{figure}
\subsection{Content Presentation}
Decluttering of SMS inbox depends largely on the way in which the final information retrieved from previous components are displayed on-device. This component is responsible for aesthetically organizing the position of key value pairs corresponding to various information extracted. Figure~\ref{fig:content} presents one way in which this module can convey a major improvement in user interface by presenting the information retrieved in an apt way.

\subsection{Pipeline Architecture}
\begin{figure}
  \centering
  \includegraphics[width=\linewidth]{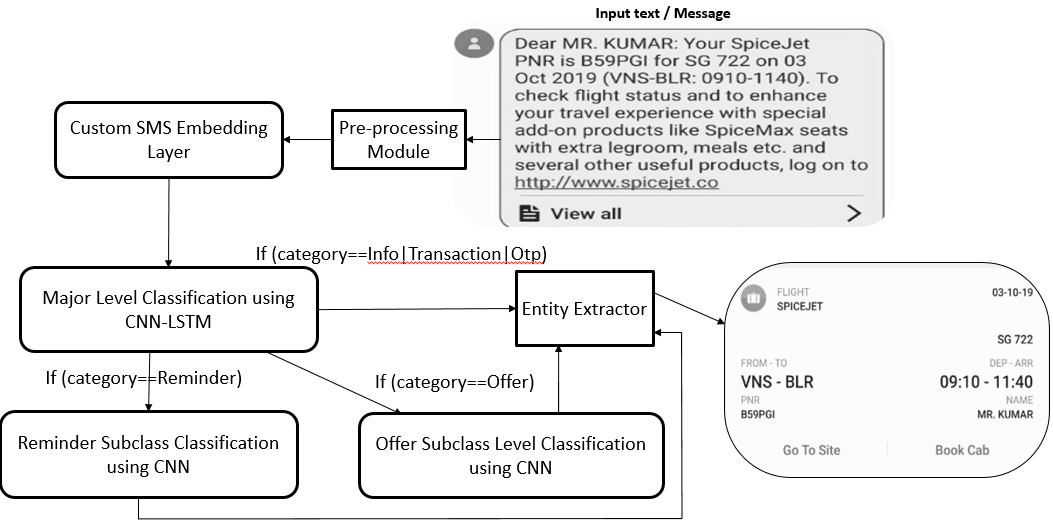}
  \caption{Pipeline Architecture}
  \label{fig:pipeline}
\end{figure}

Figure~\ref{fig:pipeline} broadly describes the functionality of the complete pipeline, which is employed for classification and content presentation of SMS.

\section{Result} \label{sec:results}

We compared different machine learning architectures in terms of their accuracy across a test data. We used the same training data as mentioned in Section~\ref{sec:corp}. We used test data of 50 SMS for each class, making the entire test data of 900 SMS. Table~\ref{tab:accuracy} shows the corresponding comparison. The prefix "All" added in model names in Table~\ref{tab:accuracy} basically implies that the same model has been used at both the levels of hierarchy. Our hybrid model performs either at par or almost equivalent to other computationally expensive models.

\begin{table}[h]
  \caption{Accuracy Comparison of Various Models}
  \label{tab:accuracy}
  \centering
  \begin{tabular}{cc}
  \hline
    \bfseries Model & \bfseries Accuracy(\%)\\
  \hline
    All-SVM & 88.87 \\
    All-CNN & 91.30 \\
    All-LSTM & 90.48 \\
    All-CNN-LSTM & 93.23\\
    Hybrid & 93.18 \\
  \hline
  \end{tabular}
\end{table}

We compared our classification results with one of the most popular application, which employs similar kind of architecture to retrieve information from SMS and displays it in a subtle way. The application, which we targeted, was SMS Organizer\footnote{SMS Organizer. \url{https://www.microsoft.com/en-us/garage/profiles/sms-organizer}}. Table~\ref{tab:comparison} captures these comparison results on 50 test SMS of each common class.

Another important factor on which we emphasized since the beginning was the on-device efficiency of our pipeline. We chose two specific metrics, which play the most important role while determining the viability of an on-device solution, inference time and model size. For on-device efficiency, even though the accuracy results were slightly better in case of All-CNN-LSTM we went ahead with a lightweight CNN model at sub-levels. As listed in Table~\ref{tab:metrics}, we showcase some of the on-device metrics for information extraction from SMS data, achieving considerably good results on a very small dataset despite employing deep learning techniques.
\begin{table}
  \caption{Comparison Results}
  \label{tab:comparison}
  \centering
  \begin{tabular}{cp{0.3\linewidth}p{0.2\linewidth}}
  \hline
    \bfseries Categories & \bfseries Hybrid Hierarchical CNN-LSTM Accuracy(\%) & \bfseries SMS Organizer Accuracy(\%) \\
  \hline
    Reminder\_Bill & 94 & 36 \\
    Reminder\_Flight & 94 & 60 \\
    Reminder\_Delivery & 94 & 92 \\
    Reminder\_Appointment & 92 & 40 \\
    Reminder\_Train & 82 & 18 \\
    Reminder\_Bus & 98 & 40 \\
    Offer\_Cab & 72 & 28 \\
    Offer\_Flight & 98 & 28 \\
    Offer\_Food & 100 & 42 \\
    Offer\_Movie & 94 & 30 \\
    Offer\_Shopping & 100 & 46 \\
    Offer\_Hotel & 78 & 24 \\
    Transaction & 94 & 94 \\
  \hline
  \end{tabular}
\end{table}
\begin{table}[h]
  \caption{On-Device Metrics}
  \label{tab:metrics}
  \centering
  \begin{tabular}{cc}
  \hline
    \bfseries Metric & \bfseries Value\\
  \hline
    Size of Hybrid Hierarchical CNN-LSTM model & 3.1 MB \\
    Average Inference Time for Classification & 39 ms \\
    Average Inference Time for Entity Extraction & 77 ms \\
    Total Average Inference Time & 116 ms \\
    Average Time taken for Information Extraction\\ of 10k SMS & 19 mins \\
  \hline
  \end{tabular}
\end{table}

\section{Conclusion \& Future Work} \label{sec:conclusion}
In this paper, we presented a unique pipeline to extract important information from SMS. This pipeline also makes on-device functionality of information extraction very efficient.  Information extraction from SMS not only prevents user from missing out on important details in messages but also helps us in resolving the cluttering of SMS inbox and of course enhancing the user experience if the entire pipeline discussed is accompanied by an aesthetically subtle user interface.

One important direction in which we can expand information extraction from SMS could be by making this pipeline language independent. Presently, we have focused only on English SMS but by using corpus of different language, we can expand the same architecture to various other languages.

\bibliographystyle{IEEEtran}
\bibliography{IEEEabrv,references}
%

\end{document}